\documentclass[letterpaper, 10 pt, conference]{ieeeconf}  

\IEEEoverridecommandlockouts                              

\usepackage{tikz}

\usepackage{xcolor}

\usepackage{multirow}
\usepackage{tabularx}
\usepackage{tabularx,booktabs}
\newcolumntype{C}{>{\centering\arraybackslash}X} 
\usepackage{ctable}
\usepackage{adjustbox}

\usepackage{graphicx}
\usepackage{mdframed}
\usepackage{tcolorbox}
\usepackage{subfigure}

\usepackage[detect-all]{siunitx}
\usepackage{amsfonts} 
\usepackage{amsmath}

\usepackage{lipsum}

\usepackage{amssymb}
\usepackage{pifont}
\newcommand{\cmark}{\ding{51}}%
\newcommand{\xmark}{\ding{55}}%

\definecolor{LightCyan}{rgb}{0.88,1,1}
\usepackage{xcolor,colortbl}

\title{\LARGE \bf
Robust 3D Object Detection in Cold Weather Conditions
}

\author{Aldi Piroli$^{1}$, Vinzenz Dallabetta$^{2}$, Marc Walessa$^{2}$, Daniel Meissner$^{2}$, \\Johannes Kopp$^{1}$, Klaus Dietmayer$^{1}$
\thanks{$^{1}$ Institute of Measurement, Control, and Microtechnology, Ulm University, Germany {\tt\small \{firstname.lastname\}@uni-ulm.de}}
\thanks{$^{2}$ BMW~AG, Petuelring 130, 80809~Munich,~Germany {\tt\small \{vinzenz.dallabetta, marc.walessa\}@bmw.de} and {\tt\small daniel.da.meissner@bmwgroup.com}}%
}

\newcommand\copyrighttext{%
	\footnotesize \copyright\,2022 IEEE. Personal use of this material is permitted. Permission from IEEE must be obtained for all other uses, in any current or future media, including reprinting/republishing this material for advertising or promotional purposes, creating new collective works, for resale or redistribution to servers or lists, or reuse of any copyrighted component of this work in other works. DOI: 10.1109/IV51971.2022.9827398}%
\newcommand\copyrightnotice{%
	\begin{tikzpicture}[remember picture,overlay]%
	\node[anchor=south,yshift=10pt] at (current page.south) {\fbox{\parbox{\dimexpr\textwidth-2cm}{\copyrighttext}}};%
	\end{tikzpicture}%
	\vspace{-10pt}%
}

\begin{document}

\maketitle
\copyrightnotice

\thispagestyle{empty}
\pagestyle{empty}

\begin{abstract}
    Adverse weather conditions can negatively affect LiDAR-based object detectors. In this work, we focus on the phenomenon of vehicle gas exhaust condensation in cold weather conditions. This everyday effect can influence the estimation of object sizes, orientations and introduce ghost object detections, compromising the reliability  of the state of the art object detectors.
    We propose to solve this problem by using data augmentation and a novel training loss term.
    
    To effectively train deep neural networks, a large set of labeled data is needed. In case of adverse weather conditions, this process can be extremely laborious and expensive. 
    We address this issue in two steps: First, we present a gas exhaust data generation method based on 3D surface reconstruction and sampling which allows us to generate large sets of gas exhaust clouds from a small pool of labeled data. 
    Second, we introduce a point cloud augmentation process that can be used to add gas exhaust to datasets recorded in good weather conditions.
    Finally, we formulate a new training loss term that leverages the augmented point cloud to increase object detection robustness by penalizing predictions that include noise.
    In contrast to other works, our method can be used with both grid-based and point-based detectors. Moreover, since our approach does not require any network architecture changes, inference times remain unchanged. 
    Experimental results on real data show that our proposed method greatly increases robustness to gas exhaust and noisy data. 
\end{abstract}

\begin{figure}[t!]
    \centering
    \includegraphics[width=\columnwidth]{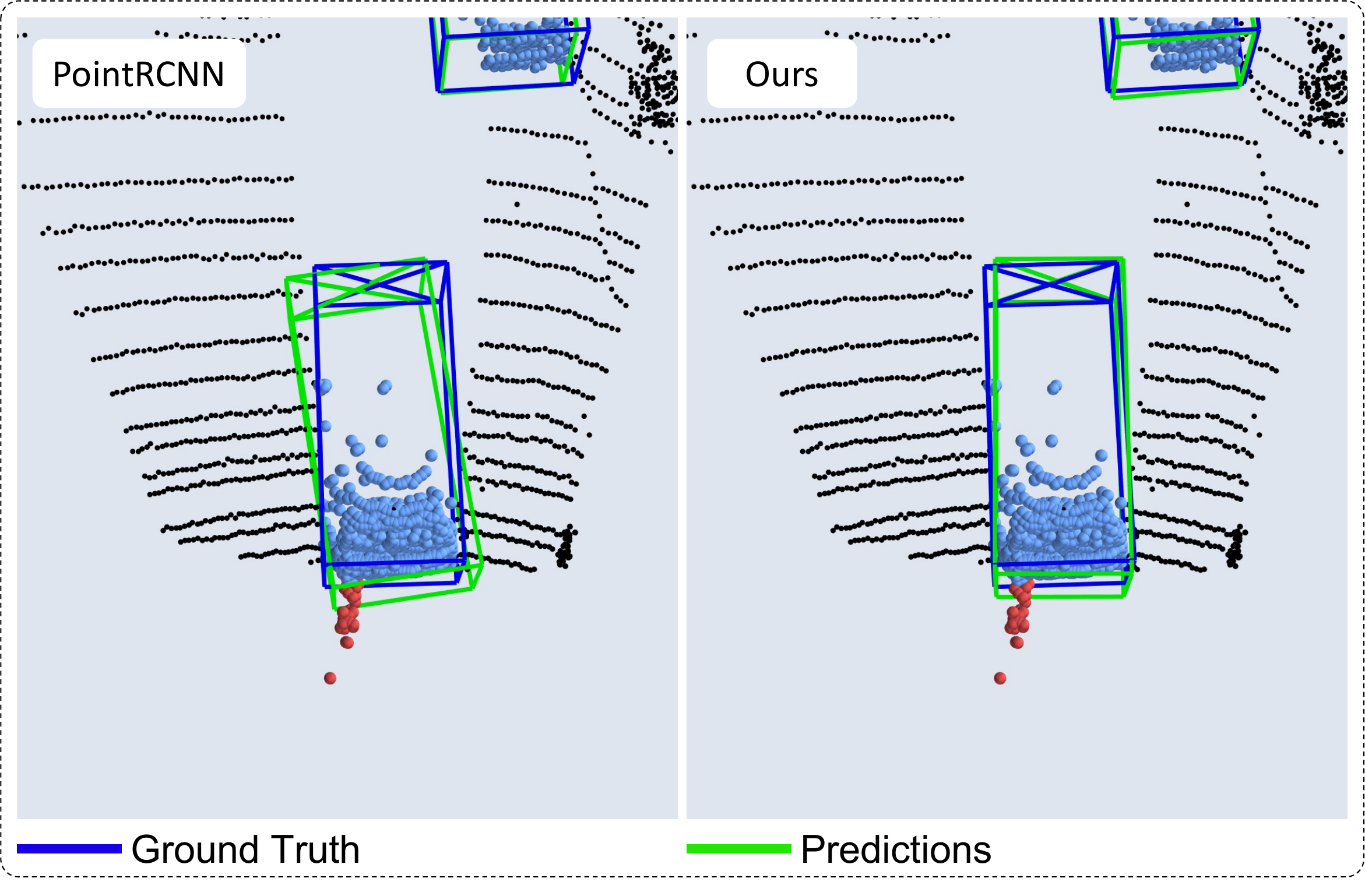}
    \caption{
    The figure above shows how vehicle gas exhaust can impact 3D object detection. 
    On the left are shown predictions derived from PointRCNN~\cite{shi2019pointrcnn}, a state of the art detector trained on the DENSE~\cite{bijelic2020seeing} dataset. On the right, the same network trained with our proposed method.
d    The baseline network predictions are heavily influenced by gas exhaust, resulting in this example in a wrong estimation of the vehicle orientation. As we can see, our method greatly improves object detection robustness when compared to the baseline network. We show in black background points, in light blue vehicles and red gas exhaust.}
    \label{Fig:teaser}
\end{figure}

\section{INTRODUCTION}
LiDARs are one of the main sensors used in autonomous driving applications, allowing for precise depth measurements independent of the lighting conditions. 
Autonomous vehicles rely on object detection to perceive and understand the surrounding environment allowing them to safely drive in complex urban scenarios.

Recent deep learning methods have allowed the training of LiDAR-based object detectors to reliably work in good weather conditions. 
It is still an open question how to achieve similar performances in adverse weather conditions since rain, snow and fog can reduce the performance of LiDAR sensors.
One other common but understudied phenomenon is vehicle gas exhaust. This occurs in cold weather conditions and is caused by the condensation of combustion engines emission plumes.
The measuring signal of a LiDAR sensor can be partially or totally reflected by the gas exhaust causing the introduction of noise in the measurements. 
Since the majority of publicly available datasets are recorded in good weather conditions, the frequency of gas exhaust instances is extremely limited. 
This makes detectors trained on such datasets sensible to its presence.   
As we can see in Fig.~\ref{Fig:teaser}, even when training on adverse weather conditions datasets~\cite{bijelic2020seeing}, the still limited number of gas exhaust training samples can lead to object detectors that wrongly estimate object sizes and orientations together with the detection of ghost objects.  
This widespread phenomenon which frequently arises in urban scenarios like traffic, red lights or pedestrian crossings, poses a challenge to the safety and reliability of upcoming autonomous vehicles.

To effectively train modern deep learning networks, a large amount of labeled data is needed. 
In case of adverse weather conditions, the process of labeling can be extremely laborious, time consuming  and expensive, due to the sparsity of point clouds and the irregularities of these effects. 
Moreover, object detectors trained in good weather conditions tend to perform poorly in adverse weather~\cite{walz2021benchmark, mirza2021robustness}. 
Finding a general strategy to increase object detection robustness without changing a network structure is not trivial due to different possible architectures that can be used. 
It is however extremely important and useful since each network design can reach state of the art results for specific tasks.

To address the problem of adverse weather data labeling, one common method consist of background subtraction labeling~\cite{heinzler2020cnn,stanislas2021airborne}, where weather effects like rain and fog are overlayed to a fixed background. 
This method is however limited to static scenes and cannot be applied to complex dynamic scenarios.
Popular driving simulators~\cite{dosovitskiy2017carla, shah2018airsim, wu2018squeezeseg} are also used to simulate adverse weather effects like rain and fog for camera images. 
This process has not yet been explored for LiDAR measurements. 
For LiDAR-only detectors~\cite{liu2020tanet} robustness is achieved by adding additional modules to a baseline network. 
Each method is however tailored to a specific architecture making it difficult to use with other object detectors. Moreover, the added complexity can considerably increase the detector's inference time.

We address the limitations mentioned above by first proposing a gas exhaust data generation method based on 3D surface reconstruction and sampling. 
We use this to generate a large set of training samples from a small pool of labeled data, thus reducing the need for manual labels.  
The generated data is used in our point cloud augmentation strategy that adds gas exhaust in datasets recorded in good weather conditions avoiding the need for new data acquisition.
Finally, we leverage our data augmentation to extend the baseline loss function of an object detector with our noise robustness loss function that penalizes predictions that include noise points. 
Differently from the state of the art, this allows to increase robustness to noise like gas exhaust without changing the object detector architecture.
The generalizability of our method is experimentally verified by testing it on SECOND~\cite{yan2018second} and PointRCNN~\cite{shi2019pointrcnn}, two popular object detectors based on fundamentally different network architectures.
We test our proposed method on the publicly available DENSE dataset~\cite{bijelic2020seeing}, which offers a diverse selection of weather conditions, allowing us to assess gas exhaust robustness on real data. Moreover, we show that our method greatly increases robustness to noisy data, reaching comparable results to the state of the art.

Our main contributions can be summarized as follows:
\begin{itemize}
    \item We propose a gas exhaust data generation method that produces large sets of data from a small pool of labeled point clouds.
    \item We propose a data augmentation strategy that can be used to expand a dataset with gas exhaust.
    \item We formulate a new training loss term that improves object detection robustness by penalizing network predictions which include noise points.
    \item We test the benefits of our method on the DENSE dataset. The results show that our method benefits both grid and point-based network architectures showing increased robustness to gas exhaust and noisy data.
\end{itemize}

\section{RELATED WORK}
\subsection{Adverse Weather Effects on LiDAR}
Adverse weather conditions like snow, rain, fog and cold weather can negatively affect LiDAR measurements~\cite{jokela2019testing, bijelic2018benchmark}.
Snow, for example, introduces random noise in the proximity of the sensor~\cite{charron2018noising}. Rain, in addition, substantially decreases the number of returned points~\cite{filgueira2017quantifying} and at high speed, the spray from leading vehicles can cause partial obstruction of the sensor field and introduce ghost objects~\cite{walz2021benchmark}.
In cold weather conditions, the water vapour in hot gas exhaust plumes of combustion vehicles condenses forming a visible gas cloud~\cite{giechaskiel2019exhaust}, which introduces noise in the measurements nearby the emitting vehicles. Multiple gas clouds can be detected when a vehicle accelerates from a standstill position~\cite{hasirlioglu2017effects}.

To alleviate the burden of manual data annotation of adverse weather points, in~\cite{heinzler2020cnn} the authors utilize a weather chamber to automatically label weather data. They first recreate a static urban scenario with stationary vehicles and mannequins for pedestrians and then overlay artificially generated fog and rain. The weather points are then automatically labeled by means of background subtraction and used to train a CNN network to classify weather points from solid points. A similar approach is used in~\cite{stanislas2021airborne} to automatically label smoke and dust.
In contrast to data labeling, synthetic data generation is also popular and is achieved using driving simulators~\cite{dosovitskiy2017carla, shah2018airsim, wu2018squeezeseg}. Although fog and rain are easily replicable for the camera view~\cite{halder2019physics}, the same methods are not directly transferable to LiDAR measurements. Moreover, the domain gap between simulated and real data is still an open research question~\cite{wu2019squeezesegv2}. Direct physics simulations of rain and fog LiDAR measurement have been proposed ~\cite{hahner2021fog, kilic2021lidar}, however, no solution has been yet offered for gas exhaust. Furthermore, a correct simulation of gas exhaust would require the modeling of aleatory forces like wind and nearby moving vehicles which affect its shape and position.

\begin{figure*}[t]
    \centering
    \includegraphics[width=\textwidth]{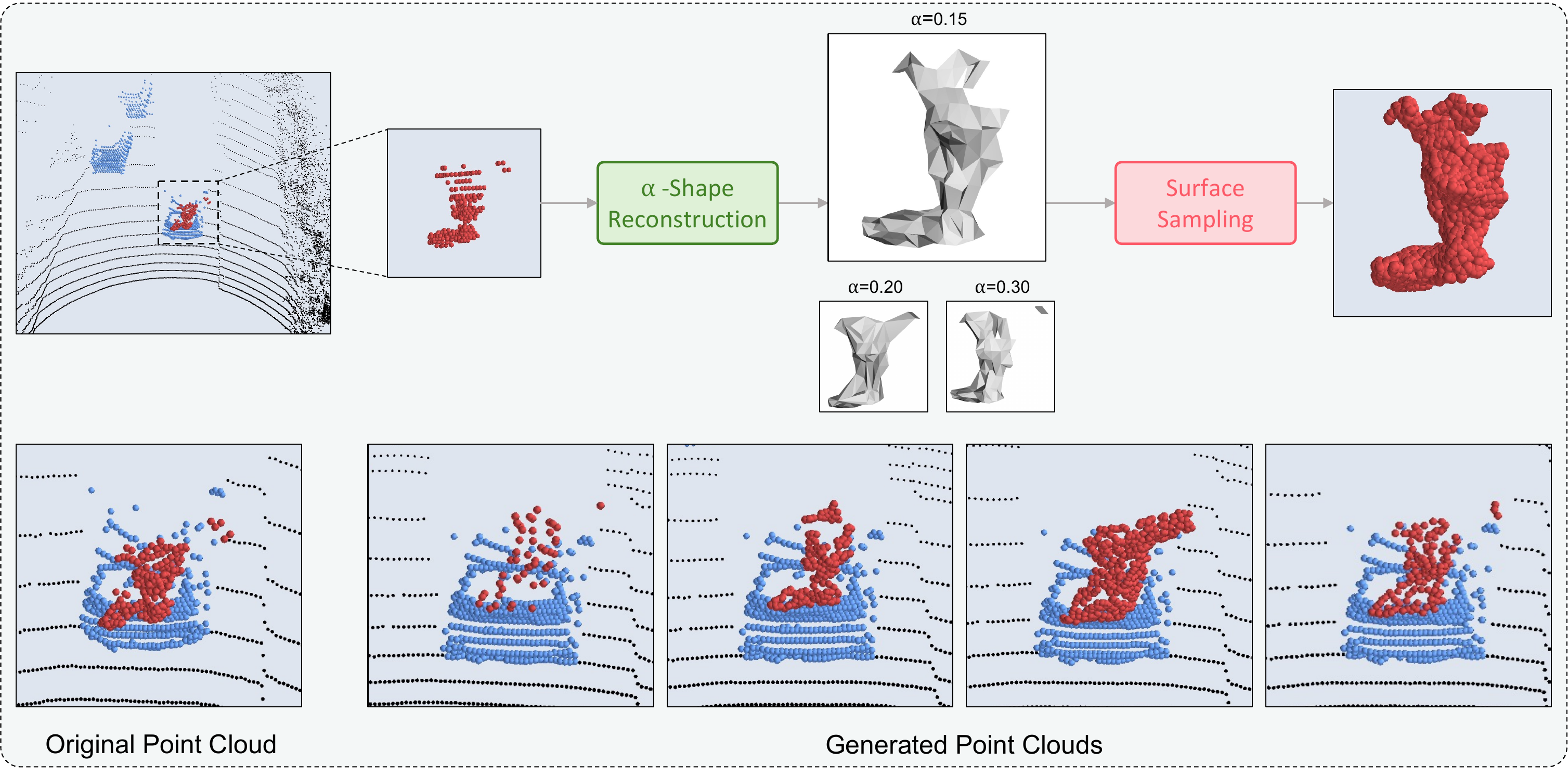}
    \caption{Overview of our proposed gas exhaust data generation. From a labeled point cloud containing gas exhaust, we reconstruct the surface of a gas cloud using the $\alpha$-Shape reconstruction algorithm with a random $\alpha$ value. From the reconstructed surface we then uniformly sample $N$ random points. On the bottom, we can see examples of point clouds generated from a single input gas exhaust cloud. The different  $\alpha$ and $N$ values allow the generation of point clouds with different shapes and densities. We show in black background points, in light blue vehicles and red gas exhaust.}
    \label{Fig:method}
\end{figure*}

\subsection{3D Object Detection in Point Clouds}
In the literature, a large body of work can be found for end-to-end deep learning on point sets. Learning can be performed by either projecting a point cloud in an intermediate representation like 2D and 3D grids or directly from the unordered points clouds using point representation networks like PointNet~\cite{qi2017pointnet}.
Modern object detectors can be divided into one-stage and two-stage methods. Single-stage detectors directly estimate the object bounding boxes without refinement allowing for fast detections. For instance, VoxelNet~\cite{zhou2018voxelnet} first projects the point cloud in a 3D voxel grid and then applies PointNet to learn voxel features that are used for 3D convolution learning. SECOND~\cite{yan2018second} improves on the design of VoxelNet by adopting 3D sparse convolutions, greatly increasing the inference speed of the network.
Two-stage detectors generate proposals for objects in the point cloud and in a second stage refine them, resulting in slower but more accurate detections.
For example, PointRCNN~\cite{shi2019pointrcnn} uses a point-based architecture to first generate high recall proposals and then refine them using a semantic segmentation mask derived from the first stage.

To increase robustness to adverse weather conditions sensor fusion architectures like~\cite{bijelic2020seeing} combine the measurements of LiDAR, radar and cameras. 
In~\cite{charron2018noising, park2020fast} classic denoising algorithms used on 2D images are used to remove snow and rain from LiDAR scans. Point wise filtering methods are however difficult to apply for effects like fog and gas exhaust where the returned measurements are dense and locally clustered.
Object detection robustness in case of multi-sensor measurements is explored in~\cite{shin2019roarnet}. This is achieved by first generating object proposals using a monocular camera which are refined using localized LiDAR measurements.
The state of the art in general data noise robustness is reached in TANet~\cite{liu2020tanet} by augmenting the single-stage detector PointPillars~\cite{lang2019pointpillars} with an additional attention based stage. This uses an attention mechanism to weight the voxelized input by combining point wise, channel wise and voxel wise information resulting in an object detector that is significantly more robust to noisy data.

\section{METHOD}
 In the following, we give a detailed description of the gas exhaust generation method together with our point cloud augmentation strategy and noise robustness loss function. Fig.~\ref{Fig:method} shows an overview of the gas exhaust generation method and an example of generated data.

\subsection{Point Sampling from 3D Shape Reconstructions }
\label{Sec:data_gen}
We assume to have a small set of labeled gas exhaust point clouds $G = \{ g_1, ..., g_n\}$  with $g_k = \{ x_k , y_k, z_k, r_k\}_{k=1}^{K}$ where $x, y, z$ are the coordinates of the points in space and $r$ the reflectivity.
We use a surface reconstruction algorithm to generate a 3D surface of a point cloud $g_i$. In our case, we use the $\alpha$-Shape algorithm~\cite{edelsbrunner1983shape} which reconstructs a 3D surface $S$ from an unstructured point cloud. Here we set the algorithm parameter $\alpha \in \mathbb{R}$ to a random value in the range $(0,1]$. This parameter can be seen as the resolution for the surface reconstruction.
From the surface $S$ we then uniformly sample $N \in \mathbb{N}$ points, with $N \in [100, 1000]$.
To recover the reflectivity of the new sampled points we find for each point its nearest neighbour belonging to the original point cloud $g_i$ and assign its reflectivity.

As we can see in the examples shown in Fig.~\ref{Fig:method}, the combination of the parameters $\alpha$ and $N$ allows us to generate a large set of gas clouds with different shapes and densities from just a single input. This substantially increases the number of available data, reducing the need for new labels.

\subsection{Point Cloud Augmentation Strategy}
\label{Sec:pcl_aug}
In the literature, one common point cloud augmentation method for unbalanced datasets is the copy-paste technique~\cite{fang2019instaboost}. 
This is used to balance the number of underrepresented classes in an unbalanced dataset by inserting objects in a point cloud from a predefined collection. 
Using the generated gas exhaust data described in the previous section, we use this same method to augment point clouds with gas exhaust points.
Given a point cloud with annotated 3D bounding boxes, for each vehicle bounding box, we add in its proximity gas exhaust with probability $p_{gas}$. The gas exhaust cloud is randomly placed either in the back center, back right corner or back left corner of the vehicle, where exhaust pipes are usually located. With probability $p_{top}$, gas exhaust is added on the top of the vehicle simulating the passing through a gas cloud emitted by a different vehicle.
To introduce more variation during training, we apply our point cloud augmentation to a LiDAR scan with probability $p_{aug}$. This probability can be set to low values at early training stages to ensure effective learning and then gradually increase it to train the network to output consistent predictions for different noise levels in the data.

A LiDAR point cloud is defined by the hardware properties of its sensor. For example, the number of laser layers together with the rotation frequency defines the maximum number of measurable points in a single scan. Due to the physical alignment of the lasers on the sensors, nearby objects tend also to have a higher number of returned points than far away ones. Therefore, simply applying our point cloud augmentation would violate the physical properties of the dataset sensor.
Similar to~\cite{yi2021complete}, we recover the original structure of the scan by first transforming each point ($x$, $y$, $z$) of the augmented point cloud into spherical coordinates ($r$, $\phi$, $\theta$), where $r=\sqrt{x^2 + y^2 + z^2}$, $\phi = atan2(\sqrt{x^2 + y^2},z)$ and $\theta = atan2(y,x)$. The point cloud is then resampled using horizontal resolution proportional to the sensor turn rate and vertical resolution proportional to the physical layers in the sensor.
With this method, we can then augment a dataset with vehicle gas exhaust avoiding the need for new data acquisition and labeling.

\subsection{Noise Robustness Loss}
\label{Sec:noise_loss}
Object detectors are trained to predict the position of objects in a point cloud by outputting 3D bounding boxes. 
Adverse weather effects like vehicle gas exhaust can affect the position of the returned bounding boxes by changing their dimensions and orientations.
Moreover, isolated gas clouds can be detected as ghost objects.
To avoid this, we design a new loss term to be used alongside the baseline detector loss function. The idea is to penalize the inclusion of noise points inside the predicted bounding boxes.

\begin{figure}[t]
    \centering
    \includegraphics[width=\columnwidth]{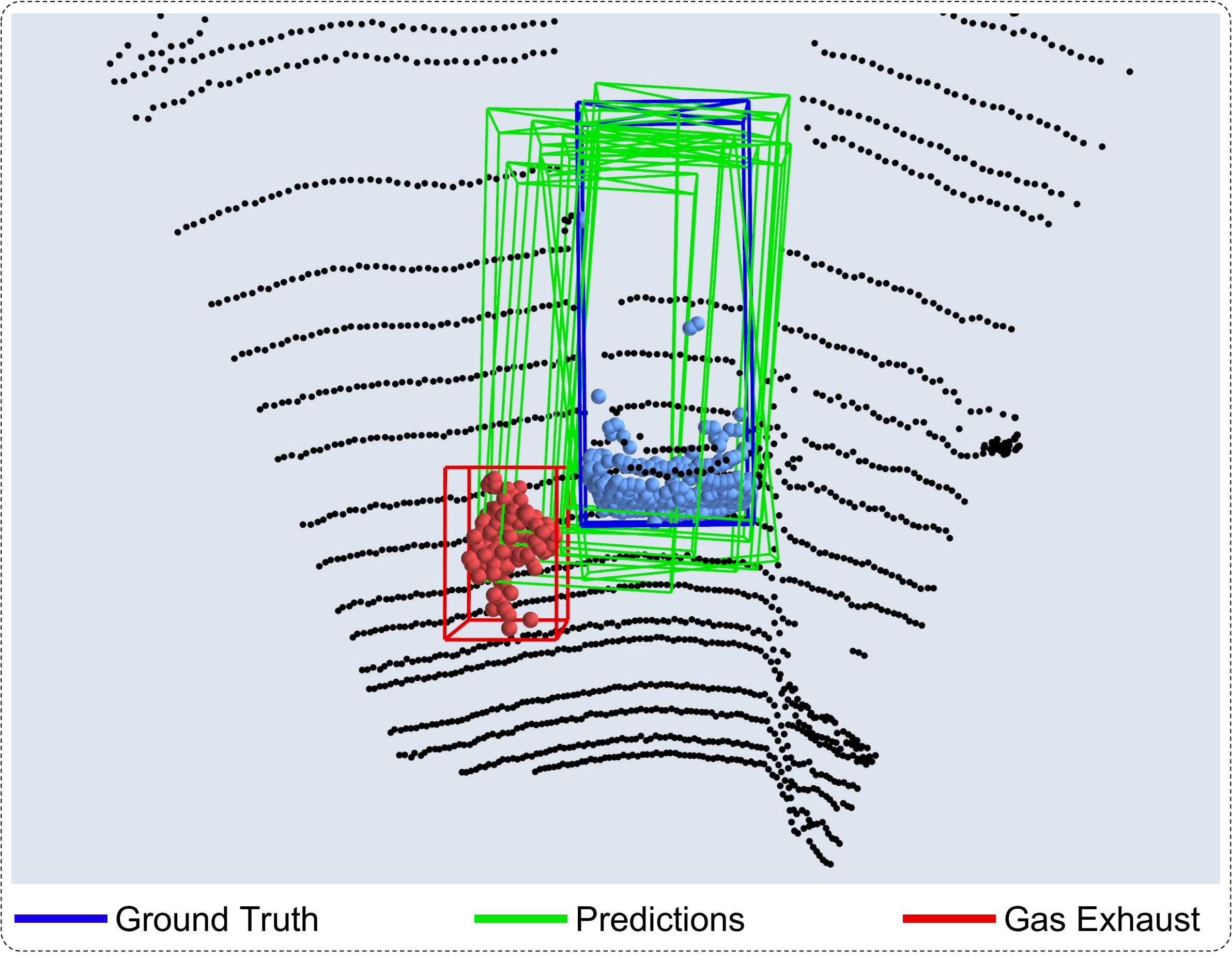}
    \caption{
 During training our noise robustness loss function computes the oriented 3D IoU between the object predictions (before non maximum suppression) and the gas exhaust boxes. 
 As shown in the figure, a large number of predictions are influenced by gas exhaust.
 Our noise robustness function penalizes such predictions during training, forcing the network to differentiate between noise and object points.
 We show in black background points, in light blue vehicles and red gas exhaust. }
    \label{Fig:loss_func}
\end{figure}

Given a point cloud augmented with the method described in Section~\ref{Sec:pcl_aug}, we extract the set of 3D bounding boxes $B$ belonging to the introduced gas exhaust points.
Our noise robustness loss can then be formulated as:
\begin{equation}
    \mathcal{L}_{\text{noise}} =  \text{IoU \textsubscript{3D} }(P,B)
\end{equation}
which expresses the oriented 3D Intersection Over Union (IoU)  loss~\cite{zhou2019iou} between the gas exhaust bounding boxes $B$ and the network object predictions $P$ before applying non maximum suppression.
The new loss term can be used alongside the object detector baseline loss function $\mathcal{L}_{\text{train}}$:
\begin{equation}
    \mathcal{L} = \mathcal{L}_{\text{train}} + \beta \mathcal{L}_{\text{noise}}
\end{equation}
with $\beta \in \mathbb{R}$ a weight parameter and $\mathcal{L}$ the total training loss. Fig.~\ref{Fig:loss_func} depicts an example of how our noise robustness loss function is used to penalize detections affected by gas exhaust. 

\section{EXPERIMENTS}
\subsection{Dataset}
For our experiments, we utilize the DENSE dataset~\cite{bijelic2020seeing}, which was recorded traveling from Germany to Sweden and contains a diverse set of weather conditions like cold weather, snow, fog and rain. We use the official training split ($3407$ scans) and test splits ($1924$ scans). To verify our proposed method on real gas exhaust scenes, we manually select all the scans from the test set  where gas exhaust is visible in the camera frame and call this gas-test set ($113$ scans).
While the gas-test set is smaller in size than the full test set, it is representative of different scenarios where gas exhaust occurs, e.g., being emitted from a leading, incoming or crossing vehicle.
We highlight that although gas exhaust is an extremely common phenomenon, only a few publicly available datasets include it, with DENSE having the largest number of instances.
We test or method using the dataset official evaluation metric, namely 3D and Bird's Eye View (BEV) Average Precision (AP) in percentage with 40 recall points across \textit{easy}, \textit{moderate} and \textit{hard} difficulty levels.
We evaluate the networks on the vehicle category, which is the main class of the DENSE dataset, using an IoU threshold of $0.7$.
For the ground truth pool of gas exhaust we manually label a set of $300$ scans of gas exhaust being emitted from a single leading vehicle. The measurements are acquired using a $40$ layers rotating LiDAR sensor which is different from the Velodyne-64 used for the DENSE dataset.

\subsection{Implementation Details}
To verify the generalizability of our method we choose two different object detectors, SECOND~\cite{yan2018second} and PointRCNN~\cite{shi2019pointrcnn}. These two networks are  representative of modern 3D object detectors since SECOND is a single-stage detector with grid-based architecture, whereas PointRCNN is a two-stage detector based on a point architecture.
Both networks are trained using the popular OpenPCDet~\cite{openpcdet2020} framework with the default training parameters. 
In the following, we will refer to them as baseline networks.
We train both networks again, using our data augmentation and noise robustness loss, but keeping the default training parameters unchanged and we call them SECOND-Aug and PointRCNN-Aug.
We also compare our results with the state of the art TANet~\cite{liu2020tanet} trained on the DENSE dataset using the official code implementation and default parameters.
The complete description of the training parameters can be found in~\cite{openpcdet2020, secondGitHub}.
We set the augmentation parameters $p_{gas} = 0.5$ and  $p_{top} = 0.1$.
During training, we initialize $p_{aug}=0$ and increase its value after each epoch by $\frac{1}{T}$, where $T$ is the total number of training epochs.
All our experiments are performed on a single NVIDIA 2080-ti GPU.

\begin{table}[t!]
    \caption{Result comparison of the baseline and augmented networks on the DENSE~\cite{bijelic2020seeing} test set. Both BEV AP and 3D AP are measured with $40$ recall points and $0.7$ IoU threshold on the vehicle class.}
    \label{Table:full_test_set}
    \centering
    \begin{adjustbox}{max width=\columnwidth}

        \begin{tabular}{l|ccc|ccc}
            \hline
            \multicolumn{1}{l|}{\multirow{2}{*}{Model}} & \multicolumn{3}{c|}{BEV AP} & \multicolumn{3}{c}{3D AP}                                                                                                        \\
            \multicolumn{1}{c|}{}                            & Easy                        & Moderate                  & Hard                      & Easy           & Moderate        & Hard                     \\ \hline
            SECOND   \cite{yan2018second}                    & \textbf{46.65}              & 43.74                     & 40.21           & \textbf{35.50} & \textbf{32.62}  & \textbf{30.38}  \\
            SECOND-Aug                                       & 46.12                       & \textbf{44.06}            & 40.21                    & 35.45          & 32.42           & 30.17                   \\
                                                             & -0.53                       & +0.31                     & 0.00                    & -0.05          & -0.20           & -0.21                \\ \hline
            PointRCNN    \cite{shi2019pointrcnn}             & \textbf{47.24}              & \textbf{45.24}           & \textbf{40.96}  & 36.47          & 34.98           & \textbf{31.22}         \\
            PointRCNN-Aug                                    & 47.12                       & 45.15                     & 40.83                    & \textbf{37.95} & \textbf{35.11} & 31.19          \\
                                                             & -0.12                       & -0.09                     & -0.13                   & +1.48          & +0.12           & -0.04                 \\ \hline
        \end{tabular}%
    \end{adjustbox}
\end{table}

\begin{table}[t!]
    \caption{Result comparison of the baseline and augmented networks on the DENSE~\cite{bijelic2020seeing} gas-test set. Both BEV AP and 3D AP are measured with $40$ recall points and $0.7$ IoU threshold on the vehicle class.}
    \label{Table:gas_exhaust_test_set}
    \centering
    \begin{adjustbox}{max width=\columnwidth}
        \begin{tabular}{l|ccc|ccc}
            \hline
            \multicolumn{1}{l|}{\multirow{2}{*}{Model}} & \multicolumn{3}{c|}{BEV AP} & \multicolumn{3}{c}{3D AP}                                                                     \\
            \multicolumn{1}{c|}{}                       & Easy                        & Moderate                  & Hard           & Easy           & Moderate       & Hard           \\ \hline
            SECOND  \cite{yan2018second}                & 62.23                       & \textbf{59.60}            & \textbf{54.60} & 56.59          & 51.62          & 46.59          \\
            SECOND-Aug                                  & \textbf{63.61}              & 59.47                     & 54.51          & \textbf{59.07} & \textbf{54.46} & \textbf{49.15} \\
                                                        & +1.38                       & -0.13                     & -0.09          & +2.48          & +2.84          & +2.56          \\ \hline
            PointRCNN    \cite{shi2019pointrcnn}        & 61.99                       & 59.51                     & 52.61          & 56.19          & 53.68          & 48.87          \\
            PointRCNN-Aug                               & \textbf{64.39}              & \textbf{60.04}            & \textbf{55.17} & \textbf{58.65} & \textbf{54.38} & \textbf{49.13} \\
                                                        & +2.40                       & +0.53                     & +2.57          & +2.46          & +0.70          & +0.26          \\ \hline
        \end{tabular}%
    \end{adjustbox}
\end{table}

\subsection{Gas Exhaust Robustness}
We start by testing both baselines and augmented networks on the full test set, which contains for the majority gas exhaust free point clouds and report the results in Table~\ref{Table:full_test_set}.
Our augmented networks yield similar results to the baseline networks, showing only a small decrease in BEV and 3D AP which can be associated with statistical noise in the data and the introduction of noise in the training data.
For PointRCNN we see however an improvement of $1.48\%$ on 3D AP \textit{easy} difficulty. 
Although the majority of the test data does not contain gas exhaust, other weather effects like snow, fog and rain are present which are attenuated as a side effect in our augmented network.
In Table~\ref{Table:gas_exhaust_test_set} the results on the gas-test set are reported.
We can see that our augmented networks are less affected by the present vehicle gas exhaust. SECOND improves up to $1.38\%$ its BEV AP and $2.84\%$ the more challenging 3D AP. 
For the \textit{moderate} and \textit{hard} difficulties, we see again a small reduction in performance that can again be associated to the same factors as in the previous experiment.
PointRCNN greatly benefits from our approach, showing a maximum improvement of $2.57\%$ for the BEV AP and $2.46\%$ for the 3D AP. 
We note that the difference in AP between the full test set and the gas-test set is due to the different number of samples in each set.

The experiment shows that our proposed method significantly improves the robustness to gas exhaust for both network architectures. 
This can be explained by the gradual introduction of gas exhaust points during training, combined with our noise robustness loss which forces the network to learn to differentiate between vehicles and nearby noise points.
Overall, the gained robustness shown by the increase in both BEV and 3D AP largely outweighs the slight reduction in performance seen in the experiments.
We emphasize that the ground truth data used for the data augmentation step was acquired using a different sensor with lower resolution than the one used for the DENSE dataset.
This shows that our augmentation method can be successfully applied for data cross over between different datasets, allowing to expand large and rich datasets with vehicle gas exhaust avoiding the need for new data acquisition and labeling.

\begin{table*}[t!]
    \caption{Data noise robustness comparison between TANet~\cite{liu2020tanet}, PointRCNN~\cite{shi2019pointrcnn} and SECOND~\cite{yan2018second} tested on the full DENSE~\cite{bijelic2020seeing} test set. We express with Aug the networks trained with our proposed method.  Both BEV AP and 3D AP are measured with $40$ recall points and $0.7$ IoU threshold on the vehicle class.}
    \label{Table:noise_robustness}
    \centering
    \begin{adjustbox}{max width=\textwidth}
        \begin{tabular}{l|c|ccc|ccc}
            \hline
            \multicolumn{1}{l|}{\multirow{2}{*}{Model}} & \multirow{2}{*}{Noise} & \multicolumn{3}{c|}{BEV AP} & \multicolumn{3}{c}{3D AP}                                                                                                                        \\
            \multicolumn{1}{c|}{}                         &                        & \multicolumn{1}{c}{Easy} & \multicolumn{1}{c}{Moderate} & \multicolumn{1}{c|}{Hard} & \multicolumn{1}{c}{Easy} & \multicolumn{1}{c}{Moderate} & \multicolumn{1}{c}{Hard} \\ \hline
            TANet   \cite{liu2020tanet}                   & \multirow{5}{*}{0}     & 65.16                       & 63.29                        & 57.40                    & 42.10                    & 40.32                    & 35.76                                                   \\ \cline{1-1} \cline{3-8}
            SECOND      \cite{yan2018second}              &                        & \textbf{46.65}              & 43.74                        & 40.21                    & \textbf{35.50}           & \textbf{32.62}           & \textbf{30.38}                                          \\
            SECOND-Aug                                    &                        & 46.12                       & \textbf{44.06}               & 40.21                    & 35.45                    & 32.42                    & 30.17                                                   \\ \cline{1-1} \cline{3-8}
            PointRCNN   \cite{shi2019pointrcnn}           &                        & \textbf{47.24}              & \textbf{45.24}               & \textbf{40.96}           & 36.47                    & 34.98                    & \textbf{31.22}                                          \\
            PointRCNN-Aug                                 &                        & 47.12                       & 45.15                        & 40.83                    & \textbf{37.95}           & \textbf{35.11}           & 31.19                                                   \\ \hline  \hline
            TANet    \cite{liu2020tanet}                  & \multirow{5}{*}{20}    & 62.22                       & 57.17                        & 54.32                    & 40.40                    & 36.14                    & 34.28                                                   \\ \cline{1-1} \cline{3-8}
            SECOND    \cite{yan2018second}                &                        & 18.50                       & 18.70                        & 17.41                    & 12.30                    & 11.99                    & 10.76                                                   \\
            SECOND-Aug                                    &                        & \textbf{41.84}              & \textbf{38.64}               & \textbf{35.04}           & \textbf{30.95}           & \textbf{27.66}           & \textbf{24.60}                                          \\ \cline{1-1} \cline{3-8}
            PointRCNN    \cite{shi2019pointrcnn}          &                        & 42.20                       & 37.94                        & 33.24                    & 32.15                    & \textbf{29.84}           & \textbf{25.81}                                          \\
            PointRCNN-Aug                                 &                        & \textbf{43.11}              & \textbf{39.38}               & \textbf{34.68}           & \textbf{32.97}           & 29.78                    & 25.59                                                   \\ \hline  \hline
            TANet   \cite{liu2020tanet}                   & \multirow{5}{*}{50}    & 59.65                       & 55.44                        & 52.26                    & 38.42                    & 35.04                    & 32.08                                                   \\ \cline{1-1} \cline{3-8}
            SECOND  \cite{yan2018second}                  &                        & 9.83                        & 9.77                         & 8.45                     & 5.91                     & 5.80                     & 5.07                                                    \\
            SECOND-Aug                                    &                        & \textbf{40.66}              & \textbf{36.97}               & \textbf{31.82}           & \textbf{28.63}           & \textbf{24.76}           & \textbf{21.00}                                          \\ \cline{1-1} \cline{3-8}
            PointRCNN  \cite{shi2019pointrcnn}            &                        & 26.00                       & 21.47                        & 18.83                    & 21.34                    & 17.53                    & 15.13                                                   \\
            PointRCNN-Aug                                 &                        & \textbf{32.49}              & \textbf{28.02}               & \textbf{23.25}           & \textbf{24.84}           & \textbf{21.14}           & \textbf{17.93}                                          \\ \hline  \hline
            TANet   \cite{liu2020tanet}                   & \multirow{5}{*}{100}   & 54.86                       & 53.16                        & 46.84                    & 34.74                    & 33.01                    & 28.50                                                   \\ \cline{1-1} \cline{3-8}
            SECOND \cite{yan2018second}                                          &                        & 4.73                        & 4.36                         & 3.93                     & 3.01                     & 2.71                     & 2.30                                                    \\
            SECOND-Aug                                    &                        & \textbf{36.49}              & \textbf{31.65}               & \textbf{26.65}           & \textbf{25.34}           & \textbf{21.29}           & \textbf{17.87}                                          \\ \cline{1-1} \cline{3-8}
            PointRCNN     \cite{shi2019pointrcnn}         &                        & 9.37                        & 7.08                         & 4.83                     & 6.47                     & 4.67                     & 4.42                                                    \\
            PointRCNN-Aug                                  &                        & \textbf{14.10}                    & \textbf{11.14}                        & \textbf{9.04}                      & \textbf{10.85}                    & \textbf{8.25}                         & \textbf{6.41}                     \\ \hline
        \end{tabular}%
    \end{adjustbox}
\end{table*}
\subsection{Data Noise Robustness}
The authors of TANet~\cite{liu2020tanet} propose to evaluate the noise robustness of an object detector by adding random noise nearby the test set ground truth bounding boxes. 
We perform the same experiment following the procedure described in their paper, specifically $k$ random points are added to each ground truth bounding box, with $k$ chosen from a uniform distribution with extremities $[0, k']$ and $k' = 0, 20, 50, 100$ to simulate different levels of noise in the point cloud.

We evaluate TANet, baseline and augmented networks on the full DENSE test set and report the results in Table~\ref{Table:noise_robustness}.
As we can see, the performance of the baseline networks quickly decreases with the increase of noise, whereas this effect is less pronounced in our augmented networks. For PointRCNN, adding a maximum of $50$ noise points results in a decrease of $16.09\%$ on the 3D AP \textit{hard} difficulty, whereas the same network trained with our proposed method decreases only by $13.26\%$~3D AP.
The results are more accentuated when increasing the number of noise points. For example, in case of a maximum of $100$ noise points, SECOND drops its BEV AP \textit{hard} difficulty performance by $90.23\%$ whereas SECOND-Aug only by $33.72\%$, showing that our proposed method greatly improves the robustness to noisy data.
We notice that in the experiment reported in~\cite{liu2020tanet}, both baseline PointRCNN and SECOND are less affected by the additional noise points than in our experiments. 
We hypothesize this to be due to the networks being trained and tested on the KITTI~\cite{geiger2013vision} dataset, that has a larger training set compared to DENSE and it is recorded only in good weather conditions. 
Our hypothesis is also supported by the reported baseline results of both networks in the case of $0$ noise points, which are substantially higher than our baselines, showing that adverse weather conditions are much more challenging for object detectors.

The drop in performance of TANet from $0$ to $100$ noise points is of $7.26\%$ 3D AP  \textit{hard} difficulty points which is lower than our two augmented networks, $12.03\%$ for SECOND and $24.78\%$ for  PointRCNN. 
This shows that the additional modules used in TANet result in a network with better noise robustness. 
We highlight however that this requires a redesign of the baseline network by adding an entire new stage, making it more difficult to integrate with other types of object detectors and increasing inference time.
In contrast, our method does not make any assumptions on the underlying structure of the network, thus making it ideal for both grid and point-based architectures. Furthermore, focusing on data augmentation and training loss allows us to keep the baseline network architecture unchanged and therefore to maintain the same inference time.

\subsection{Qualitative Results}
\begin{figure*}[t!]%
    \centerline{
    \subfigure[\centering]{\label{Fig:gas_test_set_a}{\includegraphics[width=0.24\textwidth]{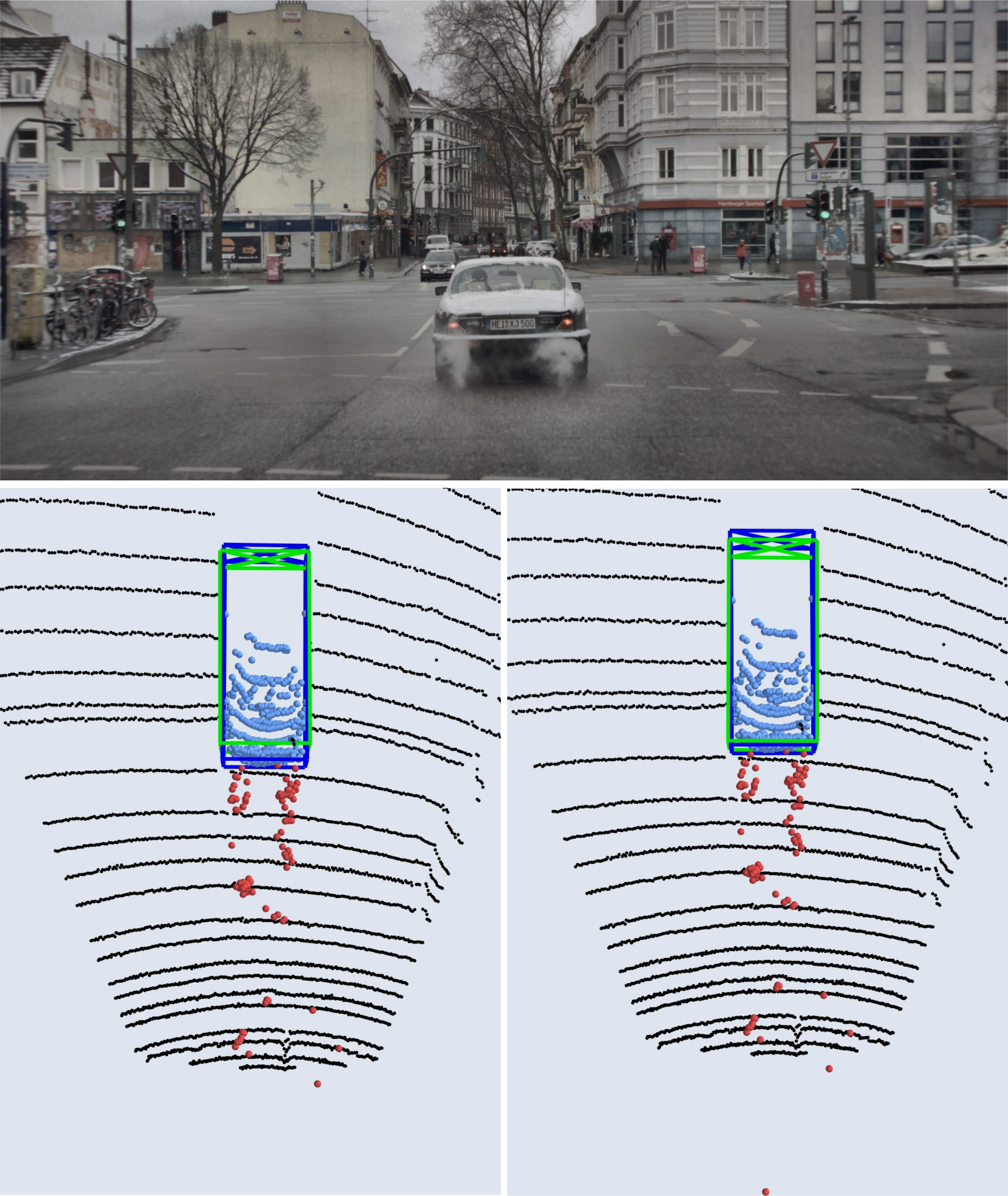} }}%
    \subfigure[\centering]{\label{Fig:gas_test_set_b}{\includegraphics[width=0.24\textwidth]{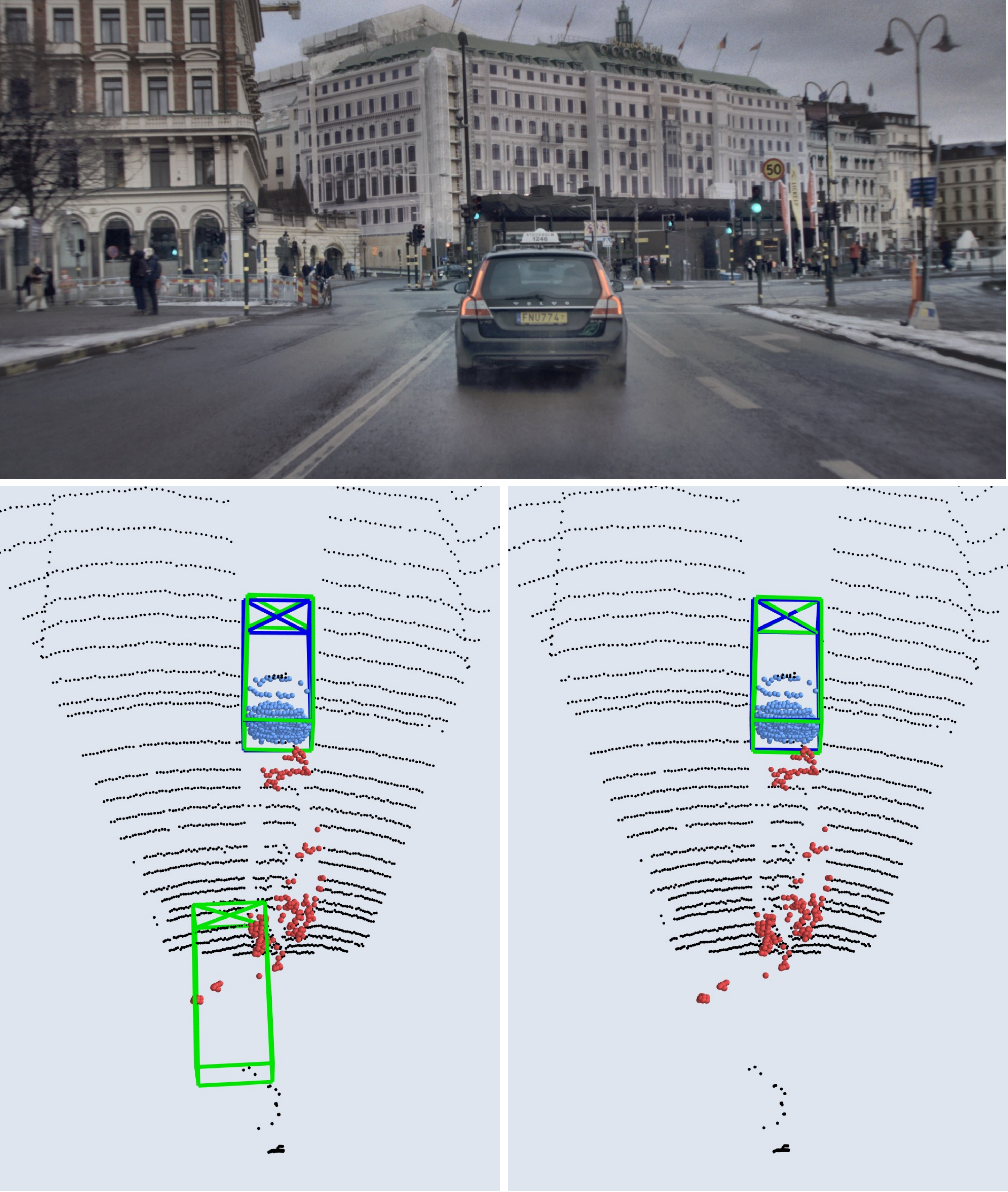} }}%
    \subfigure[\centering]{\label{Fig:gas_test_set_c}{\includegraphics[width=0.24\textwidth]{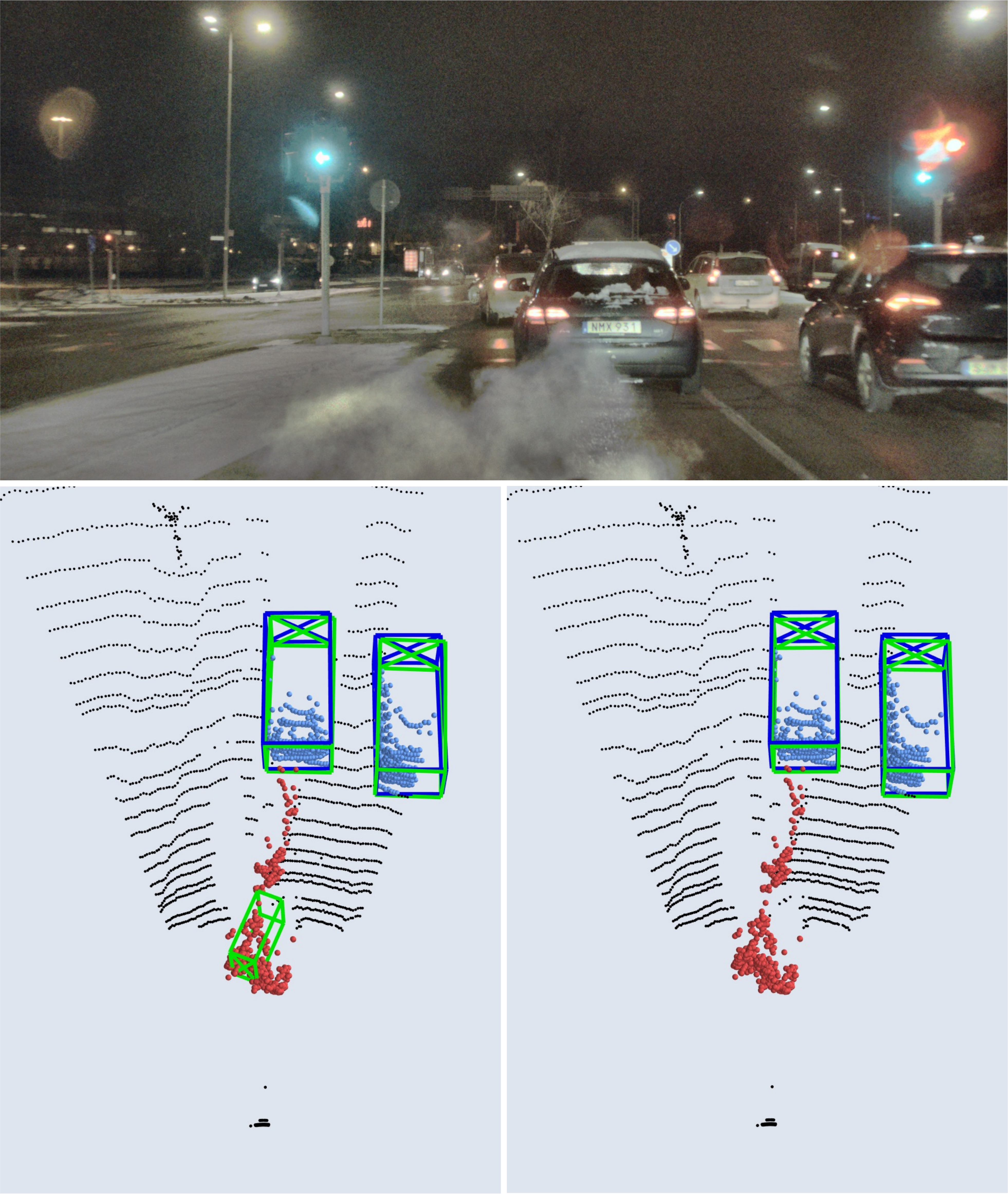} }}%
    \subfigure[\centering]{\label{Fig:gas_test_set_wrong}{\includegraphics[width=0.24\textwidth]{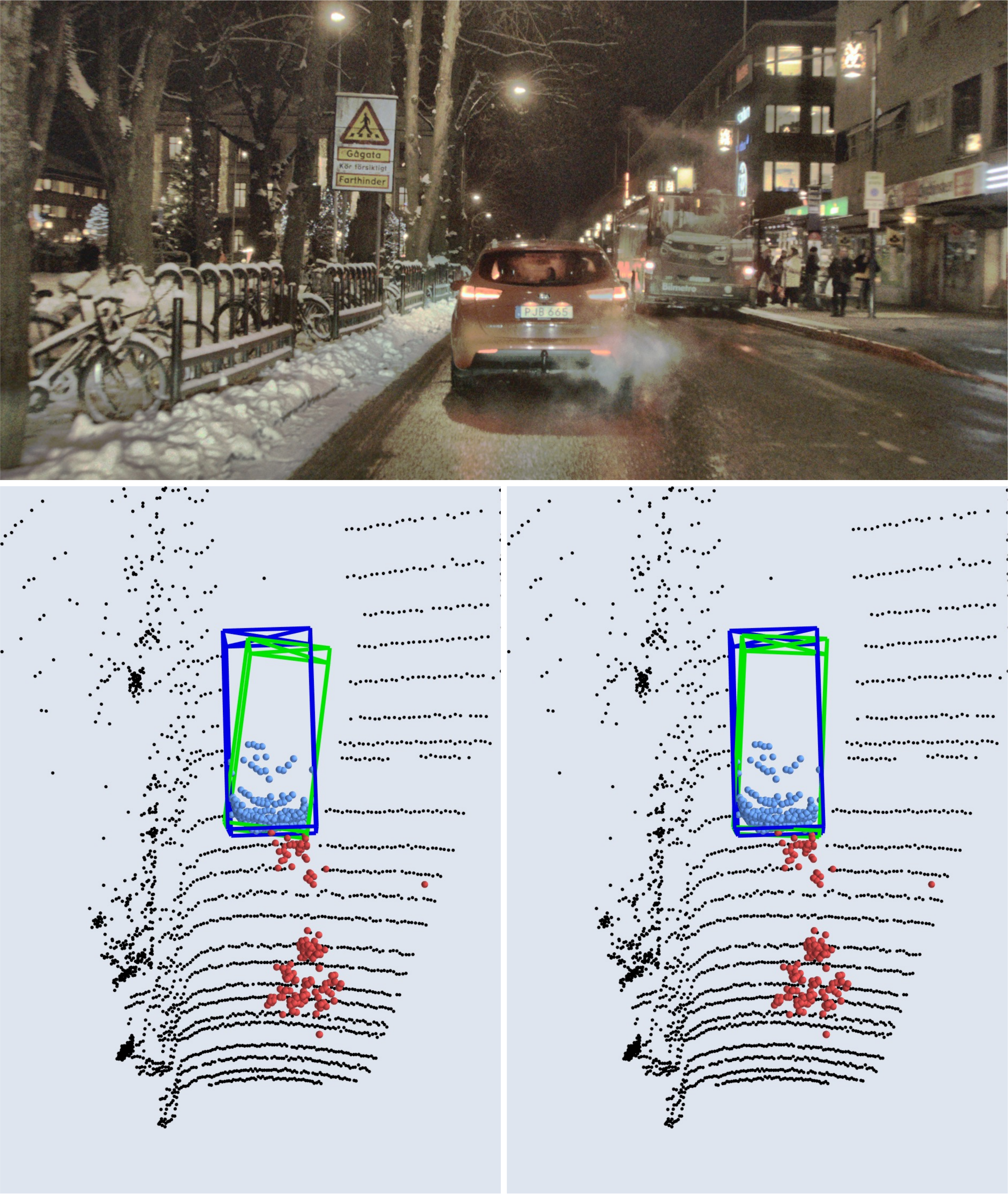} }}%
    }
    \caption{Qualitative result comparison between baseline and our augmented networks on the DENSE~\cite{bijelic2020seeing} gas-test set. In (a) and (b) we report inference from PointRCNN~\cite{shi2019pointrcnn}, whereas (c) and (d) from SECOND~\cite{yan2018second}. For each scene, on the left, we show the results of the baseline networks and on the right the same network trained with our proposed method. We can see that our method improves size and orientation estimations along with a reduction of ghost object detections. In (d) we see a failure case of our method, where the object orientation is wrongly estimated. We use blue boxes to express the ground truth and green boxes the network predictions. We show in black background points, in light blue vehicles and red gas exhaust.}%
    \label{Fig:gas_test_set}%
\end{figure*}

In Fig.~\ref{Fig:teaser} and Fig.~\ref{Fig:gas_test_set}(a), the effect of gas exhaust in the baseline network prediction is shown, where both size and orientation are influenced. In contrast, the same networks trained with our proposed method produce substantially more robust predictions.
From Fig.~\ref{Fig:gas_test_set}(b),~\ref{Fig:gas_test_set}(c), we see that some isolated gas exhaust clouds are detected as ghost objects, which could cause for example the abrupt stops of the autonomous vehicle. Instead, we see  that our proposed method greatly reduces ghost object detection caused by gas exhaust.
Fig.~\ref{Fig:gas_test_set}(d) shows an example of a failure case of our method.  We can see that both the baseline and augmented networks wrongly estimate the orientation angle of the vehicle. Such cases are often observed when the baseline network commits a large estimation error as in the example.

\section{ABLATION STUDIES}
In the following section, we discuss the influence of our design parameter choices and their effect on performance. All experiments are conducted on the DENSE gas-test set using PointRCNN as baseline network.

\subsection{Point Cloud Augmentation Effect}
We first start by analyzing the effect that our point cloud augmentation has on performance. We compare our gas exhaust generation from 3D surface sampling described in Section~\ref{Sec:data_gen} with random noise augmentation.
We sample $k$ noise points, with $k \in [100, 1000]$, from a normal distribution with $\mu=0$ and $\sigma \in (0, 0.2]$. 
The described parameters are experimentally chosen to resemble real gas exhaust clouds.
The points are then placed in a point cloud using the same strategy described in Section~\ref{Sec:pcl_aug}.
As reported in Table~\ref{Table:ablation_augmentation}, the best performance is achieved when using our augmentation method combined with our noise robustness loss function. We observe that when training  only with random noise augmented data the average performance of the baseline detector decreases by $1.02\%$~3D~AP across the three difficulty levels. 
A smaller drop in average performance of $0.36\%$~3D~AP is also seen when training only with  our augmentation method. 
These two results show that the sole introduction of noise in the training phase is detrimental to the overall performance of the network. 
Compared to our augmentation method, which generates point clouds from a pool of real data, the random noise addition might also bias the detector by introducing physically unrealizable examples during training. Moreover, as we can see in Fig.~\ref{Fig:teaser} and Fig.~\ref{Fig:gas_test_set}, the shape of gas exhaust is highly irregular and is difficult to model using a probability distribution.
This also shows that simply increasing the frequency of adverse weather phenomena like gas exhaust in the training data is not sufficient to improve robustness to such effects.
The addition of our noise robustness loss is beneficial for both augmentation methods improving the average performances of the random noise augmentation by $0.45\%$~3D~AP and the surface sampling by $1.5\%$~3D~AP across the three difficulty levels.

\subsection{Noise Loss Effect}
We test the influence of the parameter $\beta$, which weights the noise robustness loss with respect to the baseline detector loss. We can see from Table~\ref{Table:ablation_loss} that the best performance, for PointRCNN, is obtained with $\beta=0.1$, whereas for $\beta=1.0$ we have the largest drop in performance compared to the baseline.
This can be linked to the gas exhaust augmentation points placement in the point cloud, where the corresponding bounding boxes might overlap with the ground truth bounding boxes. For high values of $\beta$, the noise robustness loss can push the network predictions away from the noise bounding boxes but also from the ground truth bounding boxes.

\begin{table}[t]
    \caption{Ablation study results on the influence of our gas exhaust data augmentation compared to random noise. The results refer to PointRCNN-Aug evaluated on the DENSE~\cite{bijelic2020seeing} gas-test vehicles class.}
    \label{Table:ablation_augmentation}
    \centering
    \begin{adjustbox}{max width=\columnwidth}
        \begin{tabular}{c|c|c|cccc}
            \hline
            \multirow{2}{*}{\begin{tabular}[c]{@{}c@{}}Random \\ Noise\end{tabular}} & \multirow{2}{*}{\begin{tabular}[c]{@{}c@{}}Generated \\ Gas Exhaust\end{tabular}} & \multirow{2}{*}{\begin{tabular}[c]{@{}c@{}}Noise \\ Loss\end{tabular}} & \multicolumn{3}{c}{3D AP}                                   \\
                                                       &                                            &                                            & Easy                      & Moderate       & Hard           \\ \hline
            \xmark                                     & \xmark                                     & \xmark                                     & 56.19                     & 53.68          & 48.87          \\
            \cmark                                     & \xmark                                     & \xmark                                     & 54.76                     & 52.89          & 48.03          \\
            \xmark                                     & \cmark                                     & \xmark                                     & 55.96                     & 53.25          & 48.46          \\
            \cmark                                     & \xmark                                     & \cmark                                     & 56.30                      & 53.88          & 46.86          \\
            \xmark                                     & \cmark                                     & \cmark                                     & \textbf{58.65}            & \textbf{54.38} & \textbf{49.13} \\ \hline
        \end{tabular}%
    \end{adjustbox}
\end{table}
\begin{table}[t]
    \caption{Ablation study results on the influence of our noise robustness loss function. The results refer to PointRCNN-Aug evaluated on the DENSE~\cite{bijelic2020seeing} gas-test set vehicles class.}
    \label{Table:ablation_loss}
    \centering
    \begin{adjustbox}{max width=\columnwidth}
        \begin{tabular}{c|ccc}
            \hline
             \multirow{2}{*}{\begin{tabular}[c]{@{}c@{}}Noise Robustness \\ Loss Weight ($\beta$)\end{tabular}}  & \multicolumn{3}{c}{3D AP}                                                    \\
                                     & Easy                      & Moderate       & Hard                      \\ \hline
            0                        & 56.19                     & 53.68          & 48.87                  \\
            0.01                     & 55.94                     & 53.66          & 48.44                   \\
            0.1                      & \textbf{58.65}            & \textbf{54.38} & \textbf{49.13}  \\
            0.5                      & 56.00                     & 53.62          & 48.59                 \\
            1                        & 55.45                     & 53.14          & 48.27                  \\ \hline
        \end{tabular}%
    \end{adjustbox}
\end{table}

\section{CONCLUSION AND FUTURE WORK}
In this work, we presented a framework for increasing 3D object detection robustness against the phenomenon of vehicle gas exhaust condensation in cold weather conditions.
We first described our gas exhaust data generation method which significantly alleviates the burden of data annotation. Then, we introduced our point cloud augmentation strategy which can be used to augment datasets recorded in good weather conditions with gas exhaust avoiding the need for new data gathering and labeling.
We take advantage of our augmented data by expanding an object detector training loss with our novel noise robustness loss function which reduces the influence of noise in network predictions.
Our approach keeps the baseline object detector structure unchanged allowing it to be easily included in different network architectures without increasing inference times. 
Experimental results on the DENSE dataset show that our method increases robustness to gas exhaust and noisy data, significantly outperforming the baseline object detectors.

In future work, we will explore the use of our proposed noise robustness loss for the improvement of object detection in rainy conditions. 
\bibliographystyle{IEEEtran}
\bibliography{mybib}

\end{document}